\documentclass[sigconf]{acmart}
\usepackage{animate,graphicx,subcaption}
\usepackage{enumitem}
\usepackage{balance}
\usepackage{float}
\usepackage{multirow}
\usepackage{afterpage}
\newlength{\itemheight} %
\AtBeginDocument{%
  \providecommand\BibTeX{{%
    \normalfont B\kern-0.5em{\scshape i\kern-0.25em b}\kern-0.8em\TeX}}}

\copyrightyear{2023}
\acmYear{2023}
\setcopyright{acmlicensed}
\acmConference[MM '23] {Proceedings of the 31st ACM International Conference on Multimedia}{October 29--November 3, 2023}{Ottawa, ON, Canada.}
\acmBooktitle{Proceedings of the 31st ACM International Conference on Multimedia (MM '23), October 29--November 3, 2023, Ottawa, ON, Canada}
\acmPrice{15.00}
\acmISBN{979-8-4007-0108-5/23/10}
\acmDOI{10.1145/3581783.3612033}

\acmSubmissionID{1491}

\begin{document}

\title{Make-It-4D: Synthesizing a Consistent Long-Term Dynamic Scene Video from a Single Image}

\author{Liao Shen}
\email{leoshen@hust.edu.cn}
\affiliation{%
  \institution{School of AIA, Huazhong University
of Science and Technology}
  \country{}
  }

\author{Xingyi Li}
\email{xingyi_li@hust.edu.cn}
\affiliation{%
  \institution{School of AIA, Huazhong University
of Science and Technology}
  \country{}
  }
\affiliation{
  \institution{S-Lab, Nanyang Technological University}
  \country{}}

\author{Huiqiang Sun}
\email{shq1031@hust.edu.cn}
\affiliation{%
  \institution{School of AIA, Huazhong University
of Science and Technology}
  \country{}
  }

\author{Juewen Peng}
\email{juewenpeng@hust.edu.cn}
\affiliation{%
  \institution{School of AIA, Huazhong University
of Science and Technology}
  \country{}
  }

\author{Ke Xian}
\authornote{Corresponding author.}
\email{ke.xian@ntu.edu.sg}
\affiliation{
  \institution{S-Lab, Nanyang Technological University}
  \country{}}

\author{Zhiguo Cao}
\email{zgcao@hust.edu.cn}
\affiliation{
  \institution{School of AIA, Huazhong University
of Science and Technology}
  \country{}}

\author{Guosheng Lin}
\email{gslin@ntu.edu.sg}
\affiliation{
  \institution{S-Lab, Nanyang Technological University}
  \country{}}



\renewcommand{\shortauthors}{Liao Shen et al.}
\begin{abstract}

We study the problem of synthesizing a long-term dynamic video from only a single image. This is challenging since it requires consistent visual content movements given large camera motions. Existing methods either hallucinate inconsistent perpetual views or struggle with long camera trajectories. To address these issues, it is essential to estimate the underlying 4D (including 3D geometry and scene motion) and fill in the occluded regions. To this end, we present \textbf{Make-It-4D}, a novel method that can generate a consistent long-term dynamic video from a single image. On the one hand, we utilize layered depth images (LDIs) to represent a scene, and they are then unprojected to form a feature point cloud. To animate the visual content, the feature point cloud is displaced based on the scene flow derived from motion estimation and the corresponding camera pose. Such 4D representation enables our method to maintain the global consistency of the generated dynamic video. On the other hand, we fill in the occluded regions by using a pre-trained diffusion model to inpaint and outpaint the input image. This enables our method to work under large camera motions. Benefiting from our design, our method can be training-free which saves a significant amount of training time. Experimental results demonstrate the effectiveness of our approach, which showcases compelling rendering results. 
\end{abstract}



\maketitle
\section{Introduction}

\begin{figure}
    \setlength{\itemheight}{2.8cm}
    \begin{subfigure}{\linewidth}
        \includegraphics[height=\itemheight]{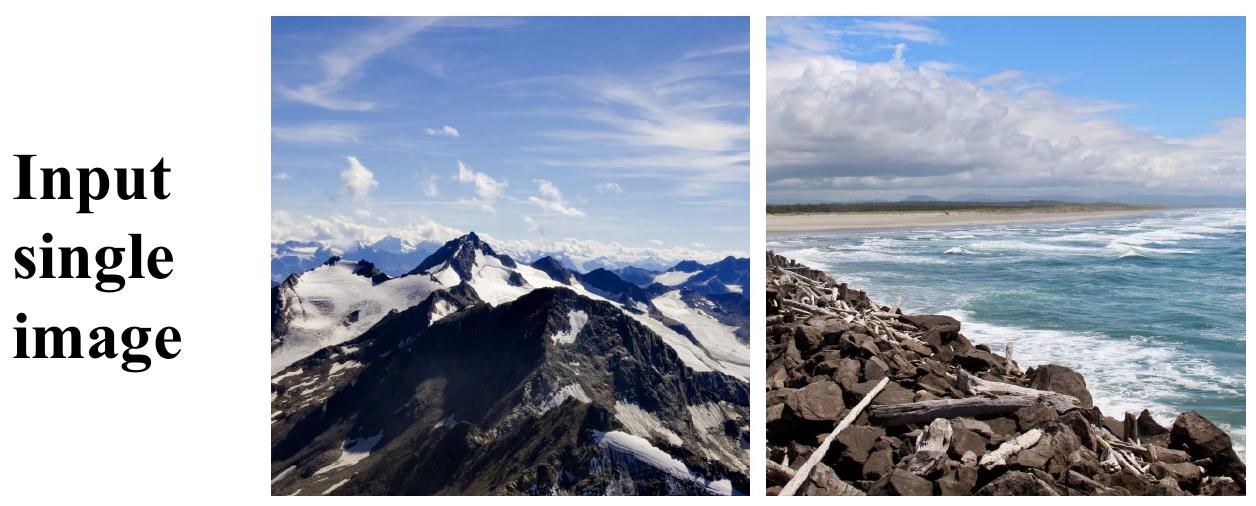}
    \end{subfigure}
    \begin{subfigure}{\linewidth}
        \vspace{0.15cm}
        \setlength\tabcolsep{0.1pt}
        \centering
        \begin{tabular}{ccc}
        
                \animategraphics[autoplay,height=\itemheight,loop,poster=last]{10}{figure/teaser/inf_65/}{0}{99} & 
                \animategraphics[autoplay,height=\itemheight,loop,poster=last]{10}{figure/teaser/cine_65/}{0}{99} &
                \animategraphics[autoplay,height=\itemheight,loop,poster=last]{10}{figure/teaser/mountain/}{0}{99}\\
                \animategraphics[autoplay,height=\itemheight,loop,poster=last]{9}{figure/teaser/inf_1076/}{0}{89} &
                \animategraphics[autoplay,height=\itemheight,loop,poster=last]{9}{figure/teaser/cine_1076/}{0}{89} &
                \animategraphics[autoplay,height=\itemheight,loop,poster=last]{9}{figure/teaser/ours_1076/}{1}{90} \\
                InfNat-zero~\cite{li2022infinitenature} & 3D Cinemagraphy~\cite{li20233d} & Ours \\
        \end{tabular}
    \end{subfigure}
    \caption{
    Given a single still image, we aim to synthesize a video featuring large camera motions and visually dynamic elements like the moving clouds and the rolling sea. Existing methods either generate inconsistent novel views~\cite{li2022infinitenature} or degrade significantly under large camera motions~\cite{li20233d}. In contrast, our method can generate a consistent long-term dynamic video. We encourage readers to view with Adobe Acrobat or KDE Okular.
    }
    \label{fig:teaser}
\end{figure}

The ubiquity of mobile cameras has made it easy for us to capture numerous photos of beautiful landscapes during our vacations. However, these static photographs only capture a single moment, lacking temporal information and failing to convey the depth and dimensionality of the scene. To create a more engaging visual experience, it is promising to explore short-form videos that allow viewers to navigate through the scene. Given a static image as in Fig.\ref{fig:teaser}, humans can imagine a dynamic scene and mentally traverse through the captured space. To achieve a similar goal, in this work, we aim to synthesize a consistent long-term dynamic video that contains both visual content movements and large camera motions. This provides an immersive and realistic experience for viewers.

Nowadays, Text-to-Video (T2V) methods ~\cite{ho2022video,ho2022imagen,singer2022make,luo2023videofusion,khachatryan2023text2video} based on the Diffusion model can generate videos from text prompts. However, they cannot control the camera pose to generate dynamic new viewpoint images and commonly require more complex diffusion model architectures and larger scale datasets. In contrast, our method constructs a 3D underlying representation of the scene, allowing the generated video to be controlled by camera trajectories.

In addition, some previous studies have focused on either synthesizing landscape flythroughs or animating a single image. 
To generate long camera track flythroughs from a single image, previous methods~\cite{liu2021infinite, li2022infinitenature, cai2022diffdreamer} continuously synthesize novel frames starting from the input image with three steps per round: render, refine, and repeat.
Specifically, they intuitively perform a per-frame refinement process to inpaint the missing regions caused by small camera motions and then repeat this process to obtain a complete camera trajectory video. Such a mechanism of using network outputs as new inputs is also prone to error accumulation, ultimately resulting in domain drifting and poor output quality. Besides, these methods cannot ensure the 3D consistency of the scene due to the lack of an underlying representation of the scene. Recently, Chai et al.~\cite{chai2023persistent} propose a method to construct an unbounded 3D world with a persistent scene representation. Yet, the resolution of the generated video is severely limited due to the high cost of volume rendering, and its 3D consistency still cannot be guaranteed. Furthermore, it is worth noting that all flythrough methods do not consider the motion of dynamic elements in the scene, such as clouds, smoke, and water. These limitations mentioned above may result in an unrealistic generated video. While single-image animation methods~\cite{endo2019animating, holynski2021animating} 
have shown promising results in producing realistic animated videos from a single image, they often face difficulties in handling camera motion.
On the other hand, 
3D Cinemagraphy~\cite{li20233d} allows for animation with slight camera motions, but it is less suitable for generating long camera trajectories. In summary, these methods suffer from the following issues: (i) perpetual view generation methods~\cite{liu2021infinite, li2022infinitenature, cai2022diffdreamer} cannot achieve global consistency since there is no explicit 3D representation; (ii) they also do not take into account the motion of dynamic elements; (iii) while 3D Cinemagraphy~\cite{li20233d} produces plausible animation of the scene, it struggles with long camera trajectories.

To address the above challenges, we present \textbf{Make-It-4D}, a novel training-free framework that generates a long-term dynamic video from a single image. To ensure the availability of complementary content when the camera moves beyond the boundaries of the initial view, we employ a pre-trained inpainting diffusion model ~\cite{rombach2022high} to outpaint the input image. To ensure 3D consistency, it is sufficient to perform a scene representation of the outpainted image, without constructing the entire 3D world. Considering this, we represent the scene as layered depth images (LDIs). 
We then utilize the inpainting diffusion model to seamlessly inpaint occluded regions of each color layer, resulting in a realistic appearance. By doing so, we eliminate the need to fill in the little holes that appear every time the camera moves, because our method has already pre-filled the information for each layer. By doing so, our method can work under large camera motions and avoid repeatedly feeding the network's output as input. This will mitigate domain drifting and inconsistency.
To animate the scene, 
we use the scene flow obtained by motion estimation to animate the point cloud. 
Since our method involves large camera motions, 
we interpolate the target camera pose to obtain intermediate camera poses for the frames in between. 
We then use these interpolated poses to render the intermediate frames, resulting in a coherent and smooth output video. 
Moreover, our framework allows for optional guidance through the use of text prompts, which can be leveraged due to the advantage of using pre-trained diffusion models.
Throughout the entire process, our framework brings still images to life, providing a vivid fly-through experience.

Extensive experiments demonstrate that our approach significantly outperforms the state-of-the-art landscape flythroughs synthesizing and 3D animation methods. We also conduct a user study to evaluate the performance of our method. 
To summarize, our key contributions include: 
\begin{itemize}[leftmargin=*]
    \item We propose \textbf{Make-It-4D}, a novel method that can synthesize a dynamic video from a single image. The generated video involves both visual content movements and large camera motions, bringing the still image back to life.
    \item We estimate the underlying 4D of the scene, including 3D scene representation and scene motion, to ensure the consistency of the generated video. To further allow large camera motions, we use a pre-trained diffusion model to inpaint and outpaint the input image to fill in the occluded regions.
    \item Our framework is entirely training-free, enabling the synthesis of long-term novel views in diverse in-the-wild scenes without the need for large-scale training.
\end{itemize}

\begin{figure*}
\begin{center}
    \includegraphics[width=1.0\linewidth]{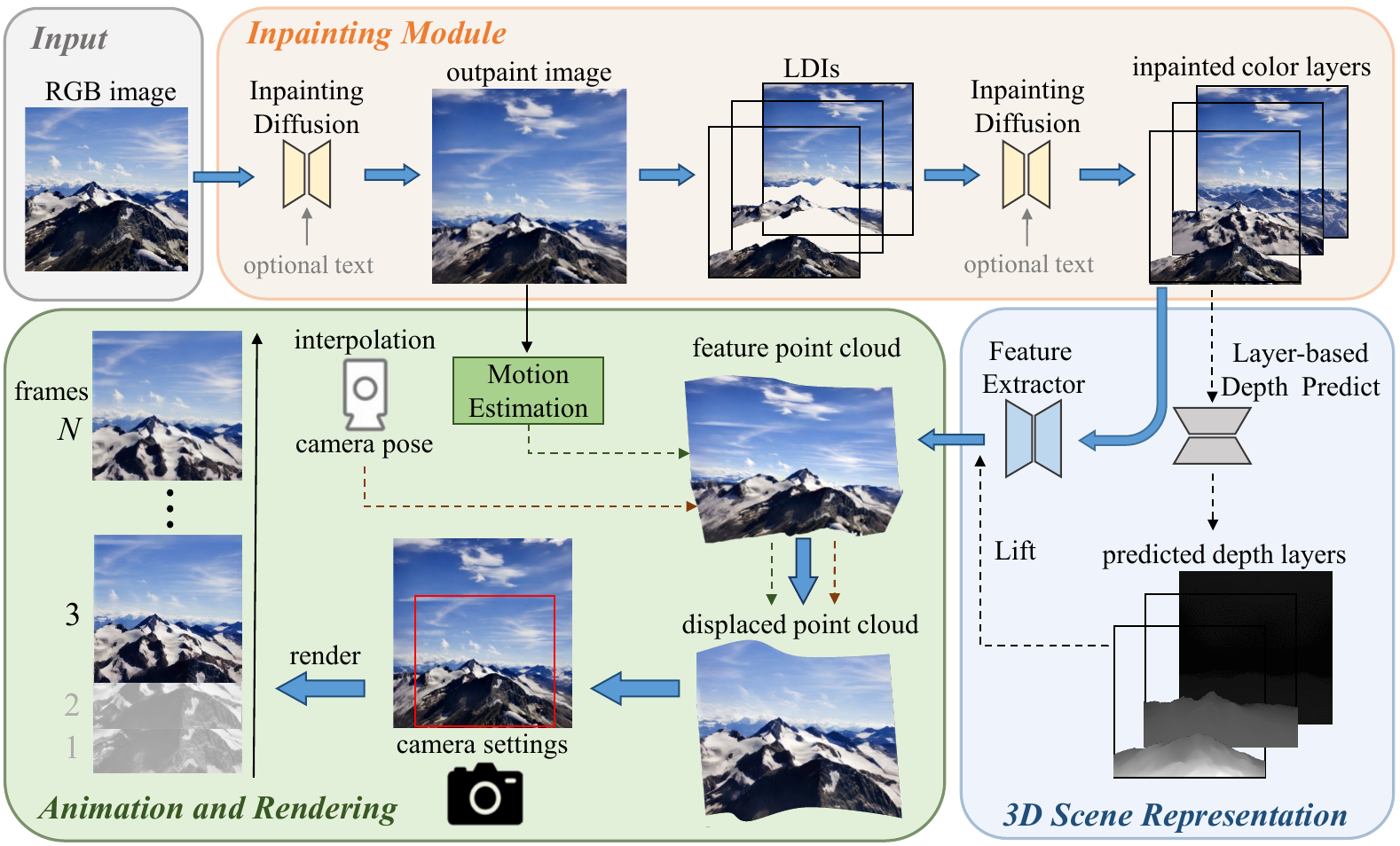}
\end{center}
    \vspace{-10pt}
   \caption{\textbf{Overview of our method.} To begin, we employ a pre-trained inpainting diffusion model to outpaint a given RGB image. 
   Next, we proceed to predict a dense depth map of the outpainted image and utilize depth discontinuities to partition the outpainted image into layered depth images (LDIs). Subsequently, we use the inpainting diffusion model to fill in obscured regions of each color layer that are obstructed by prior layers. Following this, we perform layer-based depth prediction to obtain predicted depth layers. We then utilize a 2D feature extractor to encode features from the inpainted color layers and lift them into a 3D point cloud via their corresponding predicted depth layers. The feature point cloud is displaced based on the scene flow derived from motion estimation and the corresponding camera pose. It is important to note that we set the camera intrinsics to ensure that the view size of the feature point cloud, when projected onto the target image plane, is consistent with the resolution of the initial input RGB image. To generate intermediate frames, we interpolate between the target camera pose and the initial camera pose. 
   }
\label{fig:pipeline}
\end{figure*}
\section{Related Work}

\noindent\textbf{Long-range view synthesis from a single image.}
Recent methods~\cite{liu2021infinite, li2022infinitenature, cai2022diffdreamer, koh2021pathdreamer, ren2022look, wiles2020synsin,rockwell2021pixelsynth} have proposed to synthesize scenes given a single image and camera motion as input. 
InfNat~\cite{liu2021infinite}, InfNat-Zero~\cite{li2022infinitenature}, and DiffDreamer ~\cite{cai2022diffdreamer} utilize iterative training protocols to synthesize a video depicting an in-the-wild scene captured from long camera trajectories. However, they adopt a per-frame generation
framework and lack underlying scene representations, resulting in domain drifting and inconsistent novel views. Other works~\cite{chai2023persistent, chen2023scenedreamer} propose to generate long-term views by building unbounded 3D worlds. This enables 3D consistent view generation but requires expensive computational costs. By contrast, our method only performs scene representations on the input image, 
which already inherently incorporates 3D consistency.
Moreover, we can animate the dynamic objects in the scene such as clouds and lakes while the camera moves to generate novel views, which is infeasible for all previous methods. Last but not least, these methods typically require large-scale training on videos or images from the target domain, 
whereas our approach is entirely training-free. By leveraging a pre-trained inpainting diffusion model, our method can generate novel views from a single image across diverse categories of scenes.

\noindent\textbf{Single-image animation.}
Single-image animation is the task of converting a still image into an animated video. Some works~\cite{chuang2005animating, jhou2015animating} focus on animating certain objects through physical simulations, but may not be easily applicable to the more general case of in-the-wild photos. Given videos as guidance, there are many methods that attempt to perform motion transfer on static objects~\cite{chan2019everybody, dosovitskiy2015flownet, liu2019liquid, ren2020deep, siarohin2021motion, siarohin2019first}. They require a reference video to drive the motion of static objects and thus are not suitable for our task. Therefore, we focus more on methods~\cite{endo2019animating, fan2022simulating, holynski2021animating, li2018flow, mahapatra2022controllable, li20233d} that convert still images into animated video textures by exploiting motion priors learned from a single image. For example, 3D Cinemagraphy~\cite{li20233d} jointly learns image animation and novel view synthesis in 3D. However, it struggles with large camera motions. In contrast, our method can animate dynamic objects in the image even when the camera moves over long distances.

\noindent\textbf{Text-to-3D generation.}
With the success of text-to-image generative models in recent years, text-to-3D generation has also gained a surge of interest in the community.
DreamFusion~\cite{poole2022dreamfusion} showcased impressive capability in text-to-3D synthesis by utilizing a powerful pre-trained text-to-image diffusion model~\cite{saharia2022photorealistic} as a strong image prior. 
Text-to-Video (T2V) methods ~\cite{ho2022video, ho2022imagen, singer2022make} based on diffusion model attempt to generate realistic videos from text or images. However, these methods cannot allow arbitrary camera motion.
Related to our work, SceneScape~\cite{fridman2023scenescape} utilizes a pre-trained text-to-image model to generate long videos of indoor scenes based on the input text and camera poses, but can only synthesize zoom-out trajectories and is limited in its ability to handle outdoor scenes. Instead, our method can work well in outdoor scenes and allows for arbitrary camera movement. Additionally, SceneScape requires fine-tuning of its model, while our approach is training-free.

\section{Method}
\subsection{Overview}
To the best of our knowledge, we are the first to generate consistent 
3D camera trajectories that capture the experience of flying into or out of a given RGB image, while accounting for dynamic elements in the scene such as moving clouds. At the core of Make-It-4D is a training-free framework comprising (i) an explicit 3D representation that allows our method to fly around the input image in a
consistent manner, (ii) a motion estimation module that can animate the scene, and (iii) an inpainting module that enables a larger camera motion and long camera trajectories.

We schematically illustrate our pipeline in Fig.~\ref{fig:pipeline}. Our method starts by outpainting the input RGB image 
using a pre-trained inpainting diffusion model~\cite{rombach2022high} (Sec.~\ref{sec:Inpainting}). After that, we employ a pre-trained monocular depth estimator~\cite{ranftl2021vision} to predict the depth map of the outpainted image and leverage depth discontinuities to partition the outpainted image into layered depth images (LDIs)~\cite{shade1998layered, peng2022mpib}. We then use the diffusion model to fill in the occluded regions of each color layer.
After filling in the obscured regions, we represent the scene in 3D space using layer-based depth prediction to obtain the predicted depth layers (Sec.~\ref{sec:representation}).
Next, we utilize a $2$D feature extractor to encode features from the inpainted color layers and lift them into 3D using their corresponding depth layers, resulting in a feature point cloud. The feature point cloud is then displaced based on the scene flow derived from motion estimation and the corresponding camera pose. 
We adjust the camera intrinsics to ensure that the feature point cloud, when projected onto the target image plane, matches the resolution of the initial input RGB image. Finally, we render the intermediate frames to generate a coherent video (Sec.~\ref{sec:rendering}).

\subsection{Inpainting Module}
\label{sec:Inpainting}

The information obtained from a single input image is limited. To synthesize a fly-through video, we need to obtain more information about the emerging regions outside the input field of view and the obscured regions in the input image.
3D Cinemagraphy~\cite{li20233d} appears blank regions when the camera moves out of the initial view boundaries. Besides, they use a CNN-based inpainting model from 3D Photos~\cite{shih20203d}, which can only inpaint little part of the texture and structure information.
Hence, given an input image $\mathcal{I}$, we first use the inpainting diffusion model $\mathcal{F}_{\theta_{\mathcal{F}}}$ to obtain an outpainted image $\hat{\mathcal{I}}$ that provides additional context for the scene beyond what was captured in the initial view:
\begin{equation}
    \begin{aligned}
        \hat{\mathcal{I}} = \mathcal{F}_{\theta_{\mathcal{F}}}(\mathcal{I})\, .
    \end{aligned}
\end{equation}
This approach is particularly useful when the camera moves beyond the bounds of the original view.
We also use the inpainting diffusion model to seamlessly inpaint the occluded regions. 

\subsection{Layer-based Representation}
\label{sec:representation}
\begin{figure}[!t]
  \centering
  \setlength{\belowcaptionskip}{-5pt}
  \includegraphics[width=\linewidth]{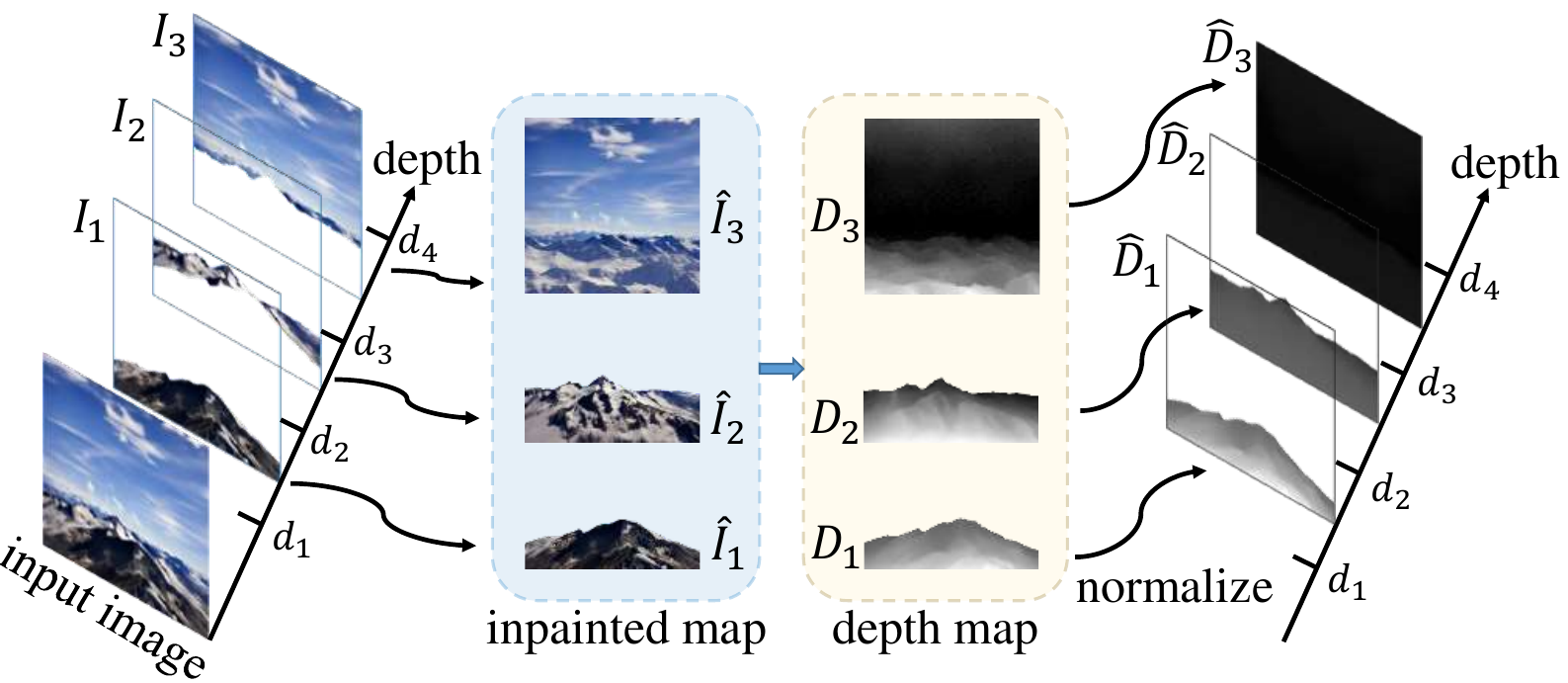}
  \vspace{-10pt}
  \caption{\textbf{Layer-based Depth predict.} We perform separate depth prediction, normalization, and remapping for each inpainted color layer, to ensure the correct depth relationship between layers.}
  \label{fig:layer-based depth}
\end{figure}

To fill in the occluded areas by diffusion model, we can intuitively refer to the previous method~\cite{liu2021infinite,li2022infinitenature,cai2022diffdreamer} that use rendering, refining, and repeating steps to inpaint missing content regions. However, since the output of the network is taken as a new input, small errors in each iteration may accumulate and eventually cause domain drifting, leading to poor output image quality. Therefore, we proposed \textit{Layer-based Depth prediction} instead of repeat operation to represent the scene, which can effectively solve the above problems and guarantee the $3$D consistency of the scene. We show the process in Fig.~\ref{fig:layer-based depth}.

\begin{figure}[!t]
  \centering
  \includegraphics[width=\linewidth]{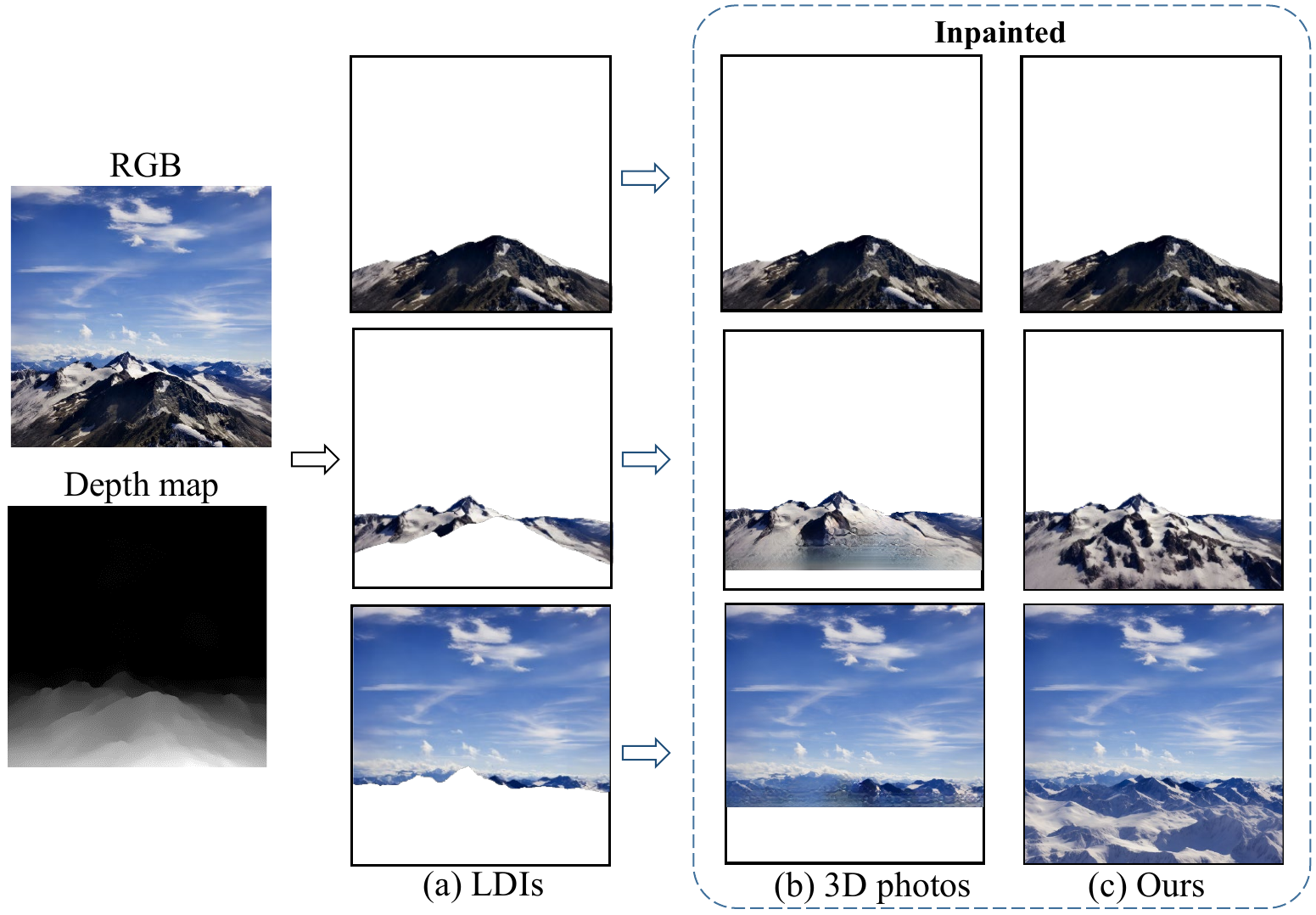}
  \vspace{-10pt}
  \caption{\textbf{From an image to inpainted RGB layers.} Given an input image and its estimated monocular depth~\cite{ranftl2021vision}, we first apply agglomerative clustering~\cite{maimon2005data} to separate the RGB image into multiple (in this case, three) RGB layers using depth map as shown in (a), then (b) 3D Photos~\cite{shih20203d} perform context-aware inpainting to obtain inpainted layers. (c) shows the results of our inpainted layers using diffusion models.}
  \label{fig:inpainted_layers}
\end{figure}

Given an outpainted image of the scene $\hat{\mathcal{I}}$, we divide the scene into different layers using depth information. To this end, we first use a pre-trained monocular depth estimation model~\cite{ranftl2021vision} denoted as $\mathcal{D}_{\theta_{\mathcal{D}}}$ to predict the depth map $\hat{D}$ of the outpainted image $\hat{\mathcal{I}}$. Then, following \cite{maimon2005data}, we use agglomerative clustering $\mathcal{C}$ to divide the depth range of the $\hat{D}$ into multiple intervals:
\begin{equation}
    \begin{aligned}
        \{ d_1, d_2, \cdots, d_i, \cdots, d_{L+1} \} = \mathcal{C}(\hat{D})\, ,
    \end{aligned}
\end{equation}
where $L$ represents the number of layers after clustering. We separate $\hat{\mathcal{I}}$ into different layers based on depth intervals, where $I_i$ denotes the i-th layered image with depth values between $d_i$ and $d_{i+1}$.
For an intermediate layer image $I_i$, it will produce many blank areas due to the occlusion of the foreground. To solve this problem, we use the diffusion model $\mathcal{F}_{\theta_{\mathcal{F}}}$ to fill these occluded areas in each layer, resulting in the inpainted layer maps $\hat{I}_i$: 
\begin{equation}
    \begin{aligned}
        \hat{I}_i = \mathcal{F}_{\theta_{\mathcal{F}}}( I_i )\, .
    \end{aligned}
\end{equation}
We compare the inpainting results obtained by different inpainting methods as shown in Fig.~\ref{fig:inpainted_layers}. One can see that the appearance of the scene obtained by the diffusion model is more realistic, which helps us ensure the $3$D consistency of the scene. 

After obtaining the inpainted layer maps, we utilize the depth estimation model $\mathcal{D}_{\theta_{\mathcal{D}}}$ to generate the depth map of each layer. It is worth noting that the inpainted color layers are not a complete image (except for the last layer), as illustrated in Fig.~\ref{fig:inpainted_layers}(c), thus we cannot directly predict the corresponding depth map. Actually, we complement the incomplete inpainted layer with colors from the layers behind it. For any inpainted layer map $\hat{I}_i$, we predict its corresponding depth map $D_i$ using the following formula: 
\begin{equation}
    \begin{aligned}
        D_i = \mathcal{D}_{\theta_{\mathcal{D}}} \left( \bigoplus_{s=i}^L \hat{I}_s \right)\, ,
    \end{aligned}
\end{equation}
where $\bigoplus$ denotes the cumulative overlay operation from the i-th layer to the L-th layer. Note that the shallower layers with lower depth values will overlay the deeper layers with higher depth values.
Since each layer uses a separate depth estimation, the original depth relationship between layers will be misaligned, resulting in incorrect $3$D representations. Therefore, we extract and normalize the depth values from all depth layers, and remap them to the value interval of the corresponding depth layer $D_i$. This process can be computed by: 
\begin{equation}
    \begin{aligned}
        \hat{D}_i = \frac{(d_{i+1}-d_i)\cdot(\hat{D}_i-min(\hat{D}_i))}{max(\hat{D}_i)-min(\hat{D}_i)}  +d_i \, ,
    \end{aligned}
\end{equation}
Through the \textit{Layer-based Depth prediction}, our method can maintain the $3$D consistency of the scene without using repeated operations.

\subsection{Scene Animation and Image Rendering}
\label{sec:rendering}
To create a $3$D-consistent representation of the scene, we first introduce a $2$D feature extraction network ~\cite{wang20223d} to encode features of inpainted color layers. Then we unproject the features into $3$D using their corresponding depth layers, resulting in a feature point cloud $\mathcal{P}=\{(X_i,f_i)\}$, where $X_i$ and $f_i$ are $3$D coordinates and the feature vector for each $3$D point respectively. To animate dynamic objects, we estimate a motion field for the observed scene. Following Holynski et al.~\cite{holynski2021animating}, we assume that a time-invariant and constant-velocity motion field, termed Eulerian flow field, can well approximate the bulk of real-world motions, such as  clouds, smoke, and water. Formally, we denote $F_{t\to t+1}(\cdot)$ as the Eulerian flow field of the scene, which represents how each pixel in the $t$ frame moves to the $t+1$ frame.
Specifically, we begin by estimating the $2$D motion field $F_{t\to t+1}(\cdot)$ from the outpainted image using a pre-trained image-to-image translation network~\cite{holynski2021animating}, and elevate this motion field into $3$D scene flow at time $t$ with the aid of estimated depth values. Then, we move each $3$D point by calculating its destination as its original position plus the scene flow. However, as points move forward, increasingly large holes can appear in the displaced point cloud. This happens when points move out of their original positions without any other points filling in those unknown regions. To address this issue, we employ the $3$D symmetric animation ~\cite{li20233d} to fill in the holes as points move forward. This allows us to obtain the final displaced point cloud $\mathcal{P}_m(t)=\{(X_i^m(t), f_i)\}$ at time $t$.

We also obtain the camera pose by adopting the autocruise algorithm from \cite{liu2021infinite}. The autocruise algorithm is particularly effective as it can adaptively adjust the camera trajectory according to the scene information. Therefore, we can use this algorithm to obtain a reasonable end-frame camera pose $c_N$ from the view of the input image $c_1$, and generate a continuous camera motion trajectory by interpolating the two camera poses to $N$ intermediate values. 

After obtaining the animated feature point clouds and camera poses, our final step is to render them into output images.
As we outpaint the initial input image to get a higher resolution and larger field of view, we adjust the camera intrinsics at each frame to ensure that the resolution of the output image is consistent with the input image.
We then use a differentiable point-based renderer ~\cite{wiles2020synsin} to splat the displaced point clouds into the target image plane, maintaining the original resolution of the input image.
As a result, we can render the displaced point clouds $\mathcal{P}_m(t)$ at time $t$ into an image.
By rendering displaced point clouds at all times, we obtain a series of rendered frames that can be compiled into the final coherent video.



\section{Experiments}

\subsection{Baselines}
\begin{figure*}[htbp]
\begin{center}
    \includegraphics[width=1.0\linewidth]{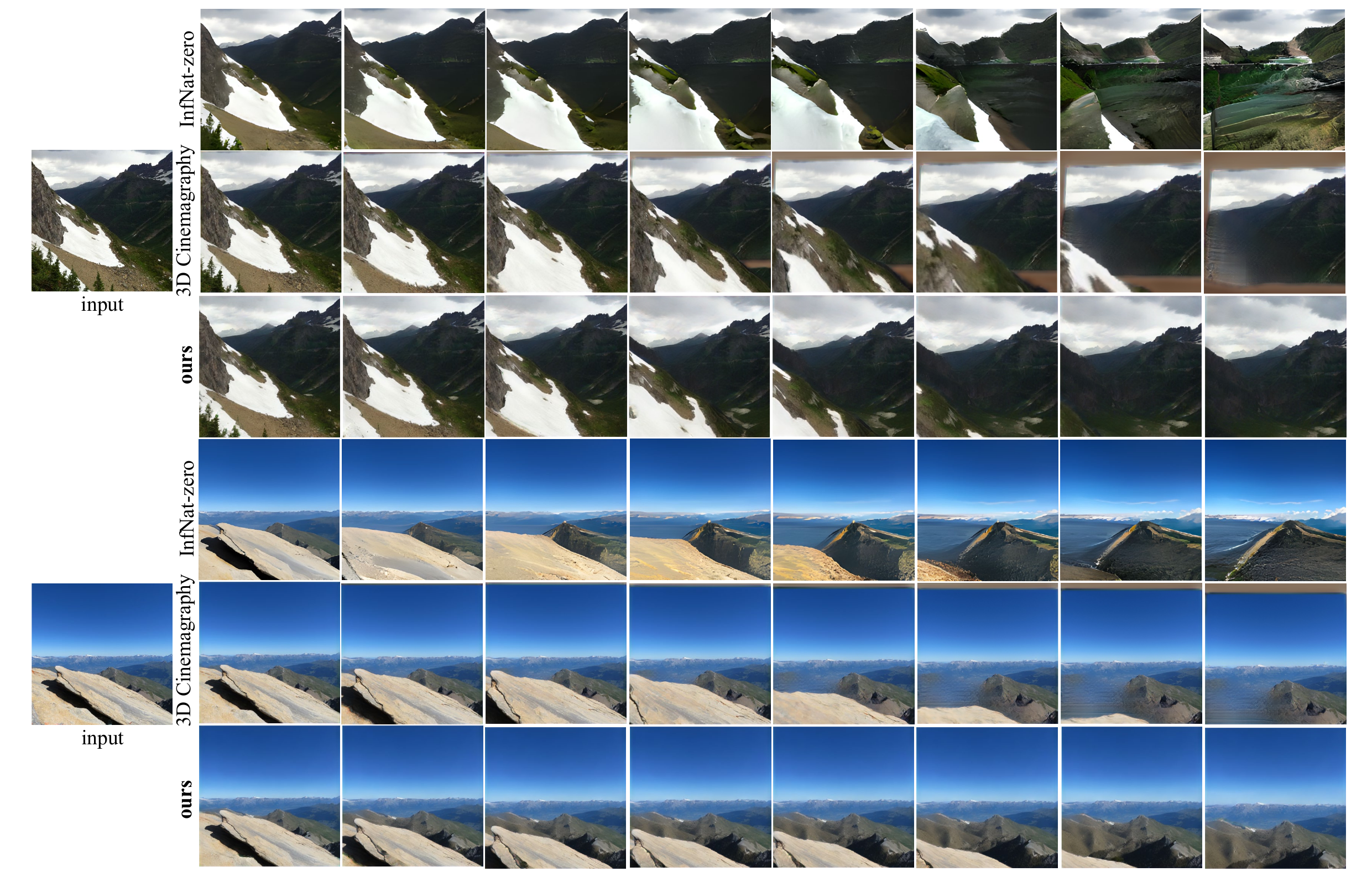}
\end{center}
    \vspace{-10pt}
   \caption{\textbf{Qualitative comparisons of baselines and our method on LHQ dataset.} From left to right, we show generated views over trajectories of length 100 for three methods: InfNat-zero ~\cite{liu2021infinite}, 3D Cinemagraphy ~\cite{li20233d}, and ours.}
\label{fig:qualitative}
\end{figure*}
To evaluate the effectiveness of our method, it is essential to compare it against the current state-of-the-art models. However, since our approach involves animation alongside long-range view generation, there are no previous works that have explored this specific combination. Therefore, a direct comparison with previous works is not possible. To overcome this challenge, we compare our method with the state-of-the-art models in the respective areas of long-range view generation and 3D animation. For long-range view generation, we use InfNat-zero~\cite{liu2021infinite}, which has achieved remarkable results in generating high-quality and diverse long-range views. For 3D animation, we use 3D Cinemagraphy~\cite{li20233d}, which can produce plausible animation of the scene while allowing slight camera movements.
As for T2V methods, they cannot achieve the goal of our work, which is to control the camera pose to generate new dynamic viewpoint images. Therefore, in our experimental settings, we cannot compare the metrics such as PSNR, SSIM, and LPIPS with them due to the inability of T2V methods to generate frames corresponding to the camera poses for comparison with the ground truth frames.

\subsection{Results}
\noindent\textbf{Evaluation dataset.} 
Following \cite{liu2021infinite}, we evaluate our method and baselines using two public datasets of nature scenes: the Landscape High Quality (LHQ) dataset~\cite{skorokhodov2021aligning}, a collection of 90K nature landscape photos, and the Aerial Coastline Imagery Dataset (ACID)~\cite{liu2021infinite}, a video dataset of nature scenes with SfM camera poses. Since there is no multi-view data for LHQ dataset, we only do qualitative experiments on it. The ACID dataset contains $279$ evaluation sequences, where each sequence has an input frame and subsequent ground truth frames and the corresponding camera pose for each frame.

\noindent\textbf{Evaluation metrics.} 
We adopt PSNR (Peak Signal to Noise Ratio), SSIM (Structural Similarity Index Measure), and Perceptual Similarity (LPIPS)~\cite{zhang2018unreasonable} as our evaluation metrics. Besides, we measure multi-view consistency using photometric error, which measures the $\mathcal{L}_1$ error when backward warping the result after one camera step to the initial frame and multiply it by 100. 
To ensure that dynamic objects' movement does not interfere with the consistency measurements, we keep the dynamic objects stationary during photometric error calculation.
All methods are evaluated at $512 \times 512$. 


\noindent\textbf{Quantitative comparisons.} We show the quantitative metrics of our method against baselines on ACID evaluation sequences in Table~\ref{tab:Quantitative comparisons}. 
Our method outperforms the other baselines in terms of view generation on all metrics, achieving highly competitive performance. 
Specifically, our approach achieves the highest PSNR and SSIM scores, which indicate that the generated views have high fidelity and are perceptually similar to the ground truth views. Moreover, our method also has the lowest LPIPS score, which indicates that our generated views are more visually similar to the ground truth views compared to the other baselines. These quantitative results demonstrate the effectiveness of our approach in generating high-quality views compared to the baseline models. 

\noindent\textbf{Qualitative comparisons.}
\begin{table}
  \caption{Quantitative comparisons on ACID evaluation sequences. Our method outperforms all baselines in all metrics, indicating that our method achieves better perceptual quality and produces a more realistic rendering.}
  \label{tab:Quantitative comparisons}
  \resizebox{1.0\linewidth}{!}{
  \renewcommand\arraystretch{1.00}
  \begin{tabular}{lccccc}
    \toprule    Method&PSNR$\uparrow$&SSIM$\uparrow$&LPIPS$\downarrow$&consistency$\downarrow$\\
    \midrule
    InfNat-zero~\cite{li2022infinitenature}   & 20.23 &0.568 &0.364 &5.36\\
    3D Cinemagraphy~\cite{li20233d}  & 21.29&0.596&0.316 &1.92\\
    Ours &\textbf{22.72} &\textbf{0.634} &\textbf{0.273} &\textbf{1.12}\\
  \bottomrule
\end{tabular}
}
\end{table}
Visual qualitative comparisons are shown in Fig. ~\ref{fig:qualitative}.
InfNat-zero quickly degenerates and leads to poor 3D consistency and semantic consistency of foreground contents due to its per-frame generation protocols and lack of underlying 3D scene representation, giving a very strong sense of unreality. Also, it cannot animate dynamic objects in the scene such as clouds. 
3D Cinemagraphy suffers from severe artifacts and voids in long-term view generation. Specifically, its rough depth inpainting causes adhesion between the different layers and leads to artifacts and holes. In the second and fifth rows, the simple texture color inpainting of 3D Cinemagraphy results in a lack of semantic information in its fills, and because it can only partially fill the image, it results in voids in the image. Moreover, the content is missing at the image boundaries after moving the camera because there is no external expansion of the image. Our method, in contrast, maintains high 3D consistency and demonstrates significantly improved synthesis quality and realism. 
We encourage readers to watch our supplementary video for a better visual comparison.

\begin{figure}[h]
  \centering
  \includegraphics[width=\linewidth]{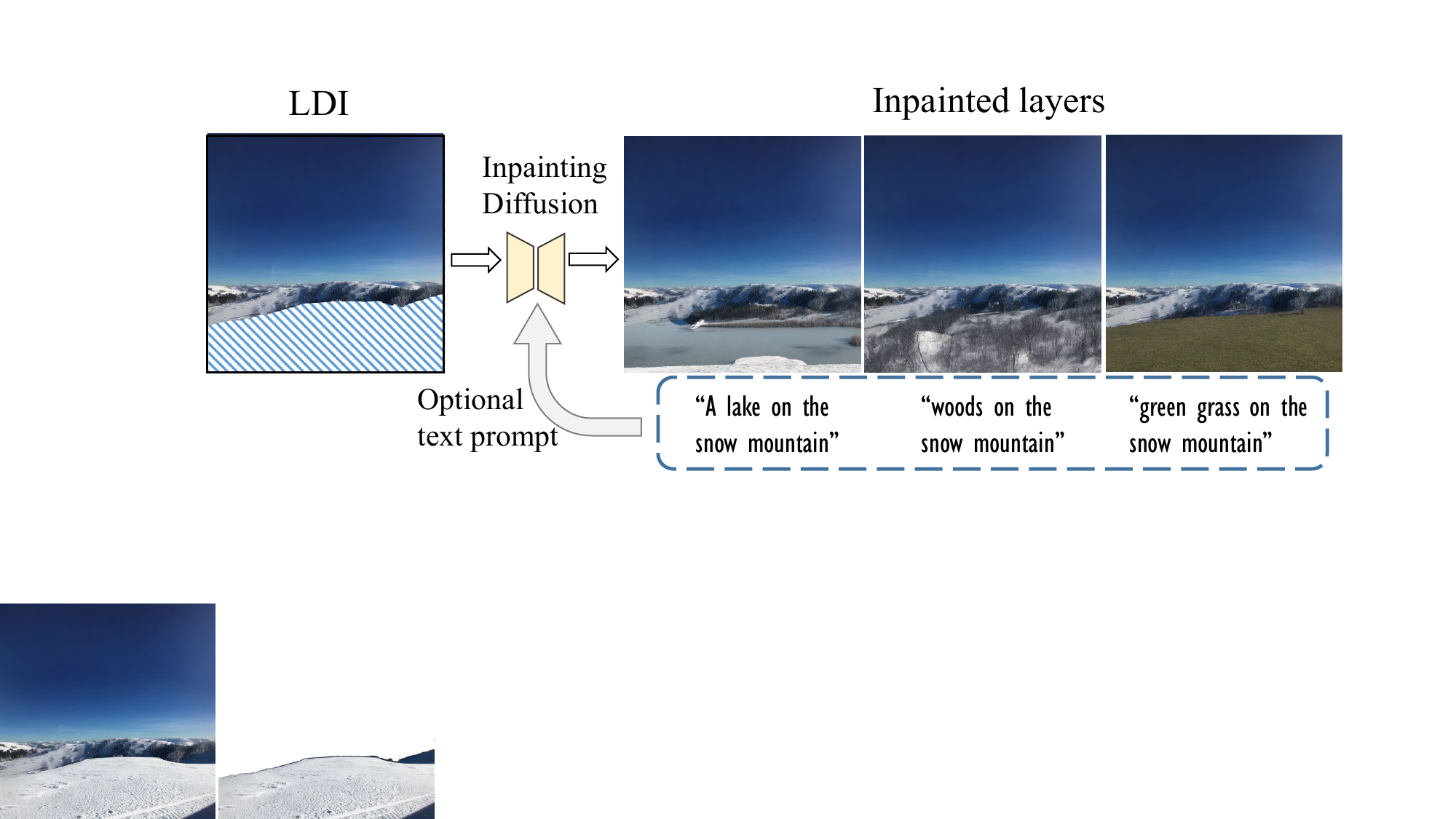}
  \caption{\textbf{Text-driven inpainting.} By using different text prompts as guidance, 
  it is possible to generate different content that can provide the desired scene experience to the user.
  }
  \label{fig:optional_inpaint}
\end{figure}

\noindent\textbf{Text driven generation.}
By using a pre-trained diffusion model that is conditioned on the user's text prompt, we can effectively use the prompt to guide the image inpainting process. Thus, we can generate diverse scenes that the users expect. Fig. ~\ref{fig:optional_inpaint} illustrates the differentiated output based on text. 
We also provide a supplementary video that demonstrates our method in more detail, showing how the text prompts can be used to generate different types of scenes and how our method produces realistic and visually appealing results. 

\begin{figure}[!t]
  \centering
  \includegraphics[width=\linewidth]{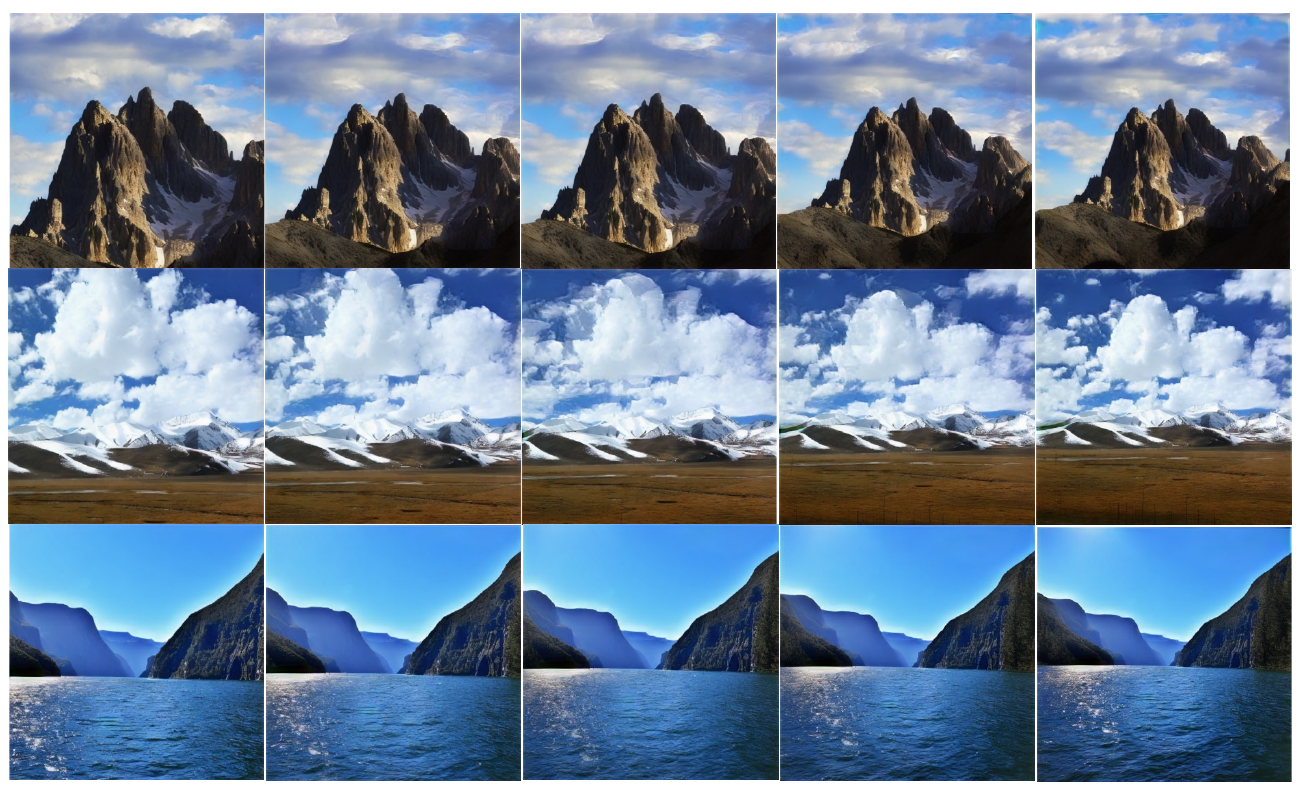}
  \caption{\textbf{Flying out of the input image.} By moving the camera backward, our method can generate the fly-out effect.}
  \label{fig:fly_out1}
\end{figure}

\noindent\textbf{Flying out of the input image.} 
Our main objective is to generate a long sequence video that simulates the visual effect of flying into an input image. Additionally, our proposed method can also generate flying-out videos by moving the camera position backward, 
as demonstrated in Fig.~\ref{fig:fly_out1}. 
This additional capability of our method enhances its versatility and usefulness in various applications. 
Readers are encouraged to view our supplementary video for better visual experiences.


\noindent\textbf{Generalization on in-the-wild photos.}
\begin{figure*}[htbp]
\begin{center}
    \includegraphics[width=1.0\linewidth]{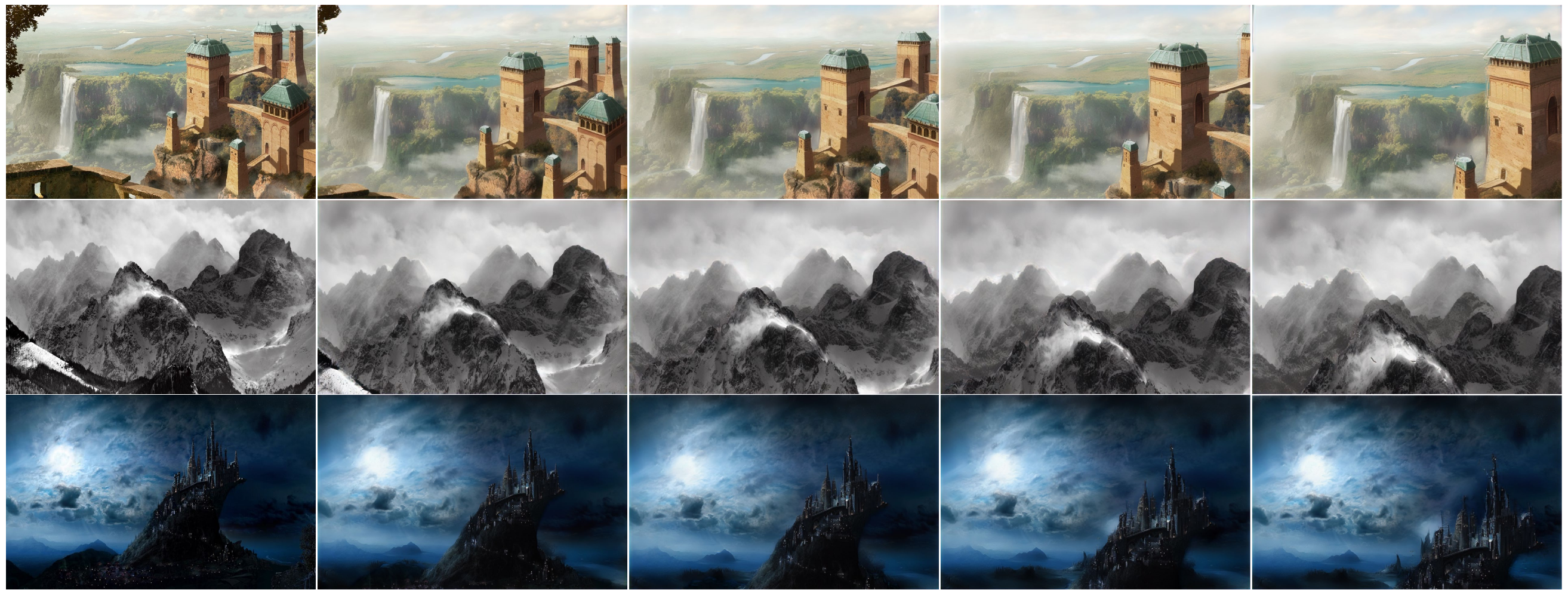}
\end{center}
    \vspace{-10pt}
   \caption{\textbf{Generalization of our method.} Our method can work at any resolution and can be generalized to paintings and historical photos in addition to real-world photos.}
\label{fig:Generalisation}
\end{figure*}
Unlike previous works that are only applicable to restricted domains, such as landscapes, due to the fact that they are trained on scenario-specific datasets, our framework can generate long-range novel views in diverse, in-the-wild scenes without training. Furthermore, existing approaches for long-term view synthesis, such as \cite{li2022infinitenature,cai2022diffdreamer}, can only render images at $128 \times 128$ resolution and require super-resolution networks to achieve higher resolution. In contrast, our method can be directly applied to any resolution.
As illustrated in Fig.~\ref{fig:Generalisation}, we present flythrough results on paintings and historical photos at $640 \times 1024$ resolution, which demonstrates the robust generalization capabilities of our approach.

\subsection{Ablation Study} 
\begin{figure}[!t]
  \centering
  \includegraphics[width=\linewidth]{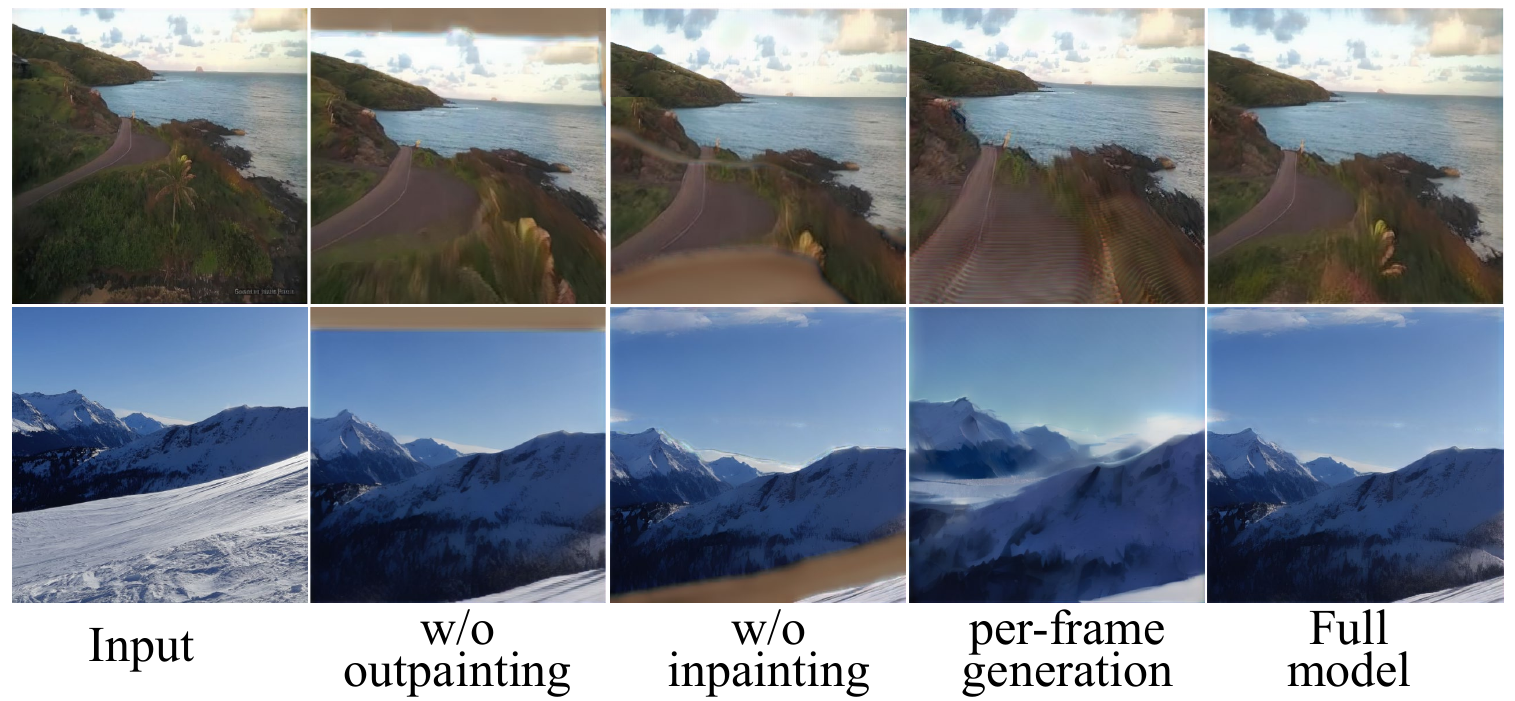}
  \vspace{-10pt}
  \caption{\textbf{Visual examples of the ablation study.} 
  Each row shows the results in different settings, presented from left to right in the following sequence: input view, results without outpainting, results without inpainting, results using per-frame generation and results obtained from our full model.
  }
  \label{fig:ablation}
\end{figure}

Each component of our system plays an important role in improving the rendering quality. To justify our design choices, we conduct ablation studies, as presented in Table~\ref{tab:ablation study}. Visual results of the ablation study are shown in Fig. ~\ref{fig:ablation}.
In the ``w/o outpainting'' experiment, 
we demonstrate that outpainting significantly improves the performance by providing realistic and reliable supplementary content when the camera moves out of the initial view boundaries. Without outpainting, there would be many holes around the image borders, as depicted in the second column of the figure.
In the ``w/o inpainting'' experiment, 
we show that if the layered depth image is not inpainted, the rendered image will exhibit holes at the depth discontinuity when the camera moves.
Finally, in the ``per-frame generation'' experiment, 
we adopt the same strategy as InfNat-zero, where we move the camera position slightly and generate one image at a time. We then use this image as the new input and repeat the process to obtain a coherent video.
As we can see, adopting such a strategy leads to poor consistency and rendering quality. 
\begin{table}[h]
  \caption{Ablation Study. Each component of our system leads to an increase in the rendering quality.}
  \vspace{-10pt}
  \label{tab:ablation study}
  \resizebox{1.0\linewidth}{!}{
  \renewcommand\arraystretch{1.00}
  \begin{tabular}{lcccc}
    \toprule
    &PSNR$\uparrow$&SSIM$\uparrow$&LPIPS$\downarrow$&Consistency$\downarrow$\\
    \midrule
     w/o outpainting  &22.29 &0.612 &0.289 &1.63\\
    w/o inpainting &21.23 &0.591 &0.335 &1.96\\
    per-frame generation &20.51 &0.577 & 0.351&3.35\\
    Full model &\textbf{22.72} &\textbf{0.634} &\textbf{0.273} &\textbf{1.12}\\
  \bottomrule
\end{tabular}
}
\end{table}

\subsection{User Study}
To better evaluate the perceptual quality and realism of our method in the view of humans, we conduct a user study to compare it with all baselines.
Specifically, we collect $30$ photos from the LHQ dataset and ACID evaluation sequences. We use different approaches to generate videos with identical settings.
During the study, participants were shown an input image and two animated videos, one generated by our method and another randomly selected approach, in random order.
107 volunteers were invited to choose the method with better perceptual quality and realism or choose none if they found it difficult to judge. Our results, presented in Table~\ref{tab:user study}, indicate that our method outperforms alternative approaches by a significant margin in terms of realism and immersion.
\begin{table}[htb]
  \caption{User study. The results indicate that our method is preferred by users as it offers a more realistic and immersive experience compared to alternative approaches.}
  \vspace{-10pt}
  \label{tab:user study}
  \resizebox{1.0\linewidth}{!}{
  \renewcommand\arraystretch{1.00}
  \begin{tabular}{lc}
    \toprule
    Comparison&Human preference\\
    \midrule
    InfNat-zero~\cite{li2022infinitenature} / \textbf{Ours}  & 3.4\%/\textbf{96.6\%}\\
    3D Cinemagraphy~\cite{li20233d} / \textbf{Ours} & 17.2\%/\textbf{82.8\%}\\
  \bottomrule
\end{tabular}
}
\end{table}

\section{Conclusion}
We present a novel approach for synthesizing long-term flythrough videos of dynamic scenes from a single image, maintaining 3D consistency, and producing realistic output videos without the need for large-scale training.
Our framework is flexible, allowing for both flying in and flying out of the input image and customization based on user-provided text prompts.
Extensive experiments demonstrate the effectiveness of our method. A user study is also conducted to validate the compelling rendering results of our method. 
We hope that our work can offer a new direction for consistent long-term dynamic scene video synthesis from a single image and inspire further research in the field.

\noindent\textbf{Limitations and future work.} 
Our approach focuses on filling in hierarchical occlusion information, but may not complement vertical information, such as obtaining a more detailed image when the camera moves forward. 
Additionally, if the depth prediction module estimates the wrong geometry from the input image, such as wrong layering, our method may not work well. 
Moreover, our method primarily focuses on handling common moving elements like fluids, while more complex object movements remain a research problem. 
We plan to explore more effective solutions to address these limitations in future work.

\begin{acks}
This work was funded by the National Natural Science Foundation of China under Grant No.U1913602 and was partly supported by the RIE2020 Industry Alignment Fund - Industry Collaboration Projects (IAF-ICP) Funding Initiative, as well as cash and in-kind contribution from the industry partner(s).
\end{acks}

\bibliographystyle{ACM-Reference-Format}
\balance
\bibliography{sample-base}


\begin{thebibliography}{40}


\ifx \showCODEN    \undefined \def \showCODEN     #1{\unskip}     \fi
\ifx \showDOI      \undefined \def \showDOI       #1{#1}\fi
\ifx \showISBNx    \undefined \def \showISBNx     #1{\unskip}     \fi
\ifx \showISBNxiii \undefined \def \showISBNxiii  #1{\unskip}     \fi
\ifx \showISSN     \undefined \def \showISSN      #1{\unskip}     \fi
\ifx \showLCCN     \undefined \def \showLCCN      #1{\unskip}     \fi
\ifx \shownote     \undefined \def \shownote      #1{#1}          \fi
\ifx \showarticletitle \undefined \def \showarticletitle #1{#1}   \fi
\ifx \showURL      \undefined \def \showURL       {\relax}        \fi
\providecommand\bibfield[2]{#2}
\providecommand\bibinfo[2]{#2}
\providecommand\natexlab[1]{#1}
\providecommand\showeprint[2][]{arXiv:#2}

\bibitem[Cai et~al\mbox{.}(2022)]%
        {cai2022diffdreamer}
\bibfield{author}{\bibinfo{person}{Shengqu Cai}, \bibinfo{person}{Eric~Ryan
  Chan}, \bibinfo{person}{Songyou Peng}, \bibinfo{person}{Mohamad Shahbazi},
  \bibinfo{person}{Anton Obukhov}, \bibinfo{person}{Luc Van~Gool}, {and}
  \bibinfo{person}{Gordon Wetzstein}.} \bibinfo{year}{2022}\natexlab{}.
\newblock \showarticletitle{DiffDreamer: Consistent Single-view Perpetual View
  Generation with Conditional Diffusion Models}.
\newblock \bibinfo{journal}{\emph{arXiv preprint arXiv:2211.12131}}
  (\bibinfo{year}{2022}).
\newblock


\bibitem[Chai et~al\mbox{.}(2023)]%
        {chai2023persistent}
\bibfield{author}{\bibinfo{person}{Lucy Chai}, \bibinfo{person}{Richard
  Tucker}, \bibinfo{person}{Zhengqi Li}, \bibinfo{person}{Phillip Isola}, {and}
  \bibinfo{person}{Noah Snavely}.} \bibinfo{year}{2023}\natexlab{}.
\newblock \showarticletitle{Persistent Nature: A Generative Model of Unbounded
  3D Worlds}.
\newblock \bibinfo{journal}{\emph{arXiv preprint arXiv:2303.13515}}
  (\bibinfo{year}{2023}).
\newblock


\bibitem[Chan et~al\mbox{.}(2019)]%
        {chan2019everybody}
\bibfield{author}{\bibinfo{person}{Caroline Chan}, \bibinfo{person}{Shiry
  Ginosar}, \bibinfo{person}{Tinghui Zhou}, {and} \bibinfo{person}{Alexei~A
  Efros}.} \bibinfo{year}{2019}\natexlab{}.
\newblock \showarticletitle{Everybody dance now}. In
  \bibinfo{booktitle}{\emph{Proceedings of the IEEE/CVF international
  conference on computer vision}}. \bibinfo{pages}{5933--5942}.
\newblock


\bibitem[Chen et~al\mbox{.}(2023)]%
        {chen2023scenedreamer}
\bibfield{author}{\bibinfo{person}{Zhaoxi Chen}, \bibinfo{person}{Guangcong
  Wang}, {and} \bibinfo{person}{Ziwei Liu}.} \bibinfo{year}{2023}\natexlab{}.
\newblock \showarticletitle{Scenedreamer: Unbounded 3d scene generation from 2d
  image collections}.
\newblock \bibinfo{journal}{\emph{arXiv preprint arXiv:2302.01330}}
  (\bibinfo{year}{2023}).
\newblock


\bibitem[Chuang et~al\mbox{.}(2005)]%
        {chuang2005animating}
\bibfield{author}{\bibinfo{person}{Yung-Yu Chuang}, \bibinfo{person}{Dan~B
  Goldman}, \bibinfo{person}{Ke~Colin Zheng}, \bibinfo{person}{Brian Curless},
  \bibinfo{person}{David~H Salesin}, {and} \bibinfo{person}{Richard Szeliski}.}
  \bibinfo{year}{2005}\natexlab{}.
\newblock \showarticletitle{Animating pictures with stochastic motion
  textures}.
\newblock In \bibinfo{booktitle}{\emph{ACM SIGGRAPH 2005 Papers}}.
  \bibinfo{pages}{853--860}.
\newblock


\bibitem[Dosovitskiy et~al\mbox{.}(2015)]%
        {dosovitskiy2015flownet}
\bibfield{author}{\bibinfo{person}{Alexey Dosovitskiy},
  \bibinfo{person}{Philipp Fischer}, \bibinfo{person}{Eddy Ilg},
  \bibinfo{person}{Philip Hausser}, \bibinfo{person}{Caner Hazirbas},
  \bibinfo{person}{Vladimir Golkov}, \bibinfo{person}{Patrick Van Der~Smagt},
  \bibinfo{person}{Daniel Cremers}, {and} \bibinfo{person}{Thomas Brox}.}
  \bibinfo{year}{2015}\natexlab{}.
\newblock \showarticletitle{Flownet: Learning optical flow with convolutional
  networks}. In \bibinfo{booktitle}{\emph{Proceedings of the IEEE international
  conference on computer vision}}. \bibinfo{pages}{2758--2766}.
\newblock


\bibitem[Endo et~al\mbox{.}(2019)]%
        {endo2019animating}
\bibfield{author}{\bibinfo{person}{Yuki Endo}, \bibinfo{person}{Yoshihiro
  Kanamori}, {and} \bibinfo{person}{Shigeru Kuriyama}.}
  \bibinfo{year}{2019}\natexlab{}.
\newblock \showarticletitle{Animating landscape: self-supervised learning of
  decoupled motion and appearance for single-image video synthesis}.
\newblock \bibinfo{journal}{\emph{arXiv preprint arXiv:1910.07192}}
  (\bibinfo{year}{2019}).
\newblock


\bibitem[Fan et~al\mbox{.}(2022)]%
        {fan2022simulating}
\bibfield{author}{\bibinfo{person}{Siming Fan}, \bibinfo{person}{Jingtan Piao},
  \bibinfo{person}{Chen Qian}, \bibinfo{person}{Kwan-Yee Lin}, {and}
  \bibinfo{person}{Hongsheng Li}.} \bibinfo{year}{2022}\natexlab{}.
\newblock \showarticletitle{Simulating Fluids in Real-World Still Images}.
\newblock \bibinfo{journal}{\emph{arXiv preprint arXiv:2204.11335}}
  (\bibinfo{year}{2022}).
\newblock


\bibitem[Fridman et~al\mbox{.}(2023)]%
        {fridman2023scenescape}
\bibfield{author}{\bibinfo{person}{Rafail Fridman}, \bibinfo{person}{Amit
  Abecasis}, \bibinfo{person}{Yoni Kasten}, {and} \bibinfo{person}{Tali
  Dekel}.} \bibinfo{year}{2023}\natexlab{}.
\newblock \showarticletitle{Scenescape: Text-driven consistent scene
  generation}.
\newblock \bibinfo{journal}{\emph{arXiv preprint arXiv:2302.01133}}
  (\bibinfo{year}{2023}).
\newblock


\bibitem[Ho et~al\mbox{.}(2022a)]%
        {ho2022imagen}
\bibfield{author}{\bibinfo{person}{Jonathan Ho}, \bibinfo{person}{William
  Chan}, \bibinfo{person}{Chitwan Saharia}, \bibinfo{person}{Jay Whang},
  \bibinfo{person}{Ruiqi Gao}, \bibinfo{person}{Alexey Gritsenko},
  \bibinfo{person}{Diederik~P Kingma}, \bibinfo{person}{Ben Poole},
  \bibinfo{person}{Mohammad Norouzi}, \bibinfo{person}{David~J Fleet},
  {et~al\mbox{.}}} \bibinfo{year}{2022}\natexlab{a}.
\newblock \showarticletitle{Imagen video: High definition video generation with
  diffusion models}.
\newblock \bibinfo{journal}{\emph{arXiv preprint arXiv:2210.02303}}
  (\bibinfo{year}{2022}).
\newblock


\bibitem[Ho et~al\mbox{.}(2022b)]%
        {ho2022video}
\bibfield{author}{\bibinfo{person}{Jonathan Ho}, \bibinfo{person}{Tim
  Salimans}, \bibinfo{person}{Alexey Gritsenko}, \bibinfo{person}{William
  Chan}, \bibinfo{person}{Mohammad Norouzi}, {and} \bibinfo{person}{David~J
  Fleet}.} \bibinfo{year}{2022}\natexlab{b}.
\newblock \showarticletitle{Video diffusion models}.
\newblock \bibinfo{journal}{\emph{arXiv preprint arXiv:2204.03458}}
  (\bibinfo{year}{2022}).
\newblock


\bibitem[Holynski et~al\mbox{.}(2021)]%
        {holynski2021animating}
\bibfield{author}{\bibinfo{person}{Aleksander Holynski},
  \bibinfo{person}{Brian~L Curless}, \bibinfo{person}{Steven~M Seitz}, {and}
  \bibinfo{person}{Richard Szeliski}.} \bibinfo{year}{2021}\natexlab{}.
\newblock \showarticletitle{Animating pictures with eulerian motion fields}. In
  \bibinfo{booktitle}{\emph{Proceedings of the IEEE/CVF Conference on Computer
  Vision and Pattern Recognition}}. \bibinfo{pages}{5810--5819}.
\newblock


\bibitem[Jhou and Cheng(2015)]%
        {jhou2015animating}
\bibfield{author}{\bibinfo{person}{Wei-Cih Jhou} {and}
  \bibinfo{person}{Wen-Huang Cheng}.} \bibinfo{year}{2015}\natexlab{}.
\newblock \showarticletitle{Animating still landscape photographs through cloud
  motion creation}.
\newblock \bibinfo{journal}{\emph{IEEE Transactions on Multimedia}}
  \bibinfo{volume}{18}, \bibinfo{number}{1} (\bibinfo{year}{2015}),
  \bibinfo{pages}{4--13}.
\newblock


\bibitem[Khachatryan et~al\mbox{.}(2023)]%
        {khachatryan2023text2video}
\bibfield{author}{\bibinfo{person}{Levon Khachatryan},
  \bibinfo{person}{Andranik Movsisyan}, \bibinfo{person}{Vahram Tadevosyan},
  \bibinfo{person}{Roberto Henschel}, \bibinfo{person}{Zhangyang Wang},
  \bibinfo{person}{Shant Navasardyan}, {and} \bibinfo{person}{Humphrey Shi}.}
  \bibinfo{year}{2023}\natexlab{}.
\newblock \showarticletitle{Text2video-zero: Text-to-image diffusion models are
  zero-shot video generators}.
\newblock \bibinfo{journal}{\emph{arXiv preprint arXiv:2303.13439}}
  (\bibinfo{year}{2023}).
\newblock


\bibitem[Koh et~al\mbox{.}(2021)]%
        {koh2021pathdreamer}
\bibfield{author}{\bibinfo{person}{Jing~Yu Koh}, \bibinfo{person}{Honglak Lee},
  \bibinfo{person}{Yinfei Yang}, \bibinfo{person}{Jason Baldridge}, {and}
  \bibinfo{person}{Peter Anderson}.} \bibinfo{year}{2021}\natexlab{}.
\newblock \showarticletitle{Pathdreamer: A world model for indoor navigation}.
  In \bibinfo{booktitle}{\emph{Proceedings of the IEEE/CVF International
  Conference on Computer Vision}}. \bibinfo{pages}{14738--14748}.
\newblock


\bibitem[Li et~al\mbox{.}(2023)]%
        {li20233d}
\bibfield{author}{\bibinfo{person}{Xingyi Li}, \bibinfo{person}{Zhiguo Cao},
  \bibinfo{person}{Huiqiang Sun}, \bibinfo{person}{Jianming Zhang},
  \bibinfo{person}{Ke Xian}, {and} \bibinfo{person}{Guosheng Lin}.}
  \bibinfo{year}{2023}\natexlab{}.
\newblock \showarticletitle{3D Cinemagraphy from a Single Image}. In
  \bibinfo{booktitle}{\emph{Proceedings of the IEEE/CVF Conference on Computer
  Vision and Pattern Recognition}}. \bibinfo{pages}{4595--4605}.
\newblock


\bibitem[Li et~al\mbox{.}(2018)]%
        {li2018flow}
\bibfield{author}{\bibinfo{person}{Yijun Li}, \bibinfo{person}{Chen Fang},
  \bibinfo{person}{Jimei Yang}, \bibinfo{person}{Zhaowen Wang},
  \bibinfo{person}{Xin Lu}, {and} \bibinfo{person}{Ming-Hsuan Yang}.}
  \bibinfo{year}{2018}\natexlab{}.
\newblock \showarticletitle{Flow-grounded spatial-temporal video prediction
  from still images}. In \bibinfo{booktitle}{\emph{Proceedings of the European
  Conference on Computer Vision (ECCV)}}. \bibinfo{pages}{600--615}.
\newblock


\bibitem[Li et~al\mbox{.}(2022)]%
        {li2022infinitenature}
\bibfield{author}{\bibinfo{person}{Zhengqi Li}, \bibinfo{person}{Qianqian
  Wang}, \bibinfo{person}{Noah Snavely}, {and} \bibinfo{person}{Angjoo
  Kanazawa}.} \bibinfo{year}{2022}\natexlab{}.
\newblock \showarticletitle{Infinitenature-zero: Learning perpetual view
  generation of natural scenes from single images}. In
  \bibinfo{booktitle}{\emph{Computer Vision--ECCV 2022: 17th European
  Conference, Tel Aviv, Israel, October 23--27, 2022, Proceedings, Part I}}.
  Springer, \bibinfo{pages}{515--534}.
\newblock


\bibitem[Liu et~al\mbox{.}(2021)]%
        {liu2021infinite}
\bibfield{author}{\bibinfo{person}{Andrew Liu}, \bibinfo{person}{Richard
  Tucker}, \bibinfo{person}{Varun Jampani}, \bibinfo{person}{Ameesh Makadia},
  \bibinfo{person}{Noah Snavely}, {and} \bibinfo{person}{Angjoo Kanazawa}.}
  \bibinfo{year}{2021}\natexlab{}.
\newblock \showarticletitle{Infinite nature: Perpetual view generation of
  natural scenes from a single image}. In \bibinfo{booktitle}{\emph{Proceedings
  of the IEEE/CVF International Conference on Computer Vision}}.
  \bibinfo{pages}{14458--14467}.
\newblock


\bibitem[Liu et~al\mbox{.}(2019)]%
        {liu2019liquid}
\bibfield{author}{\bibinfo{person}{Wen Liu}, \bibinfo{person}{Zhixin Piao},
  \bibinfo{person}{Jie Min}, \bibinfo{person}{Wenhan Luo}, \bibinfo{person}{Lin
  Ma}, {and} \bibinfo{person}{Shenghua Gao}.} \bibinfo{year}{2019}\natexlab{}.
\newblock \showarticletitle{Liquid warping gan: A unified framework for human
  motion imitation, appearance transfer and novel view synthesis}. In
  \bibinfo{booktitle}{\emph{Proceedings of the IEEE/CVF International
  Conference on Computer Vision}}. \bibinfo{pages}{5904--5913}.
\newblock


\bibitem[Luo et~al\mbox{.}(2023)]%
        {luo2023videofusion}
\bibfield{author}{\bibinfo{person}{Zhengxiong Luo}, \bibinfo{person}{Dayou
  Chen}, \bibinfo{person}{Yingya Zhang}, \bibinfo{person}{Yan Huang},
  \bibinfo{person}{Liang Wang}, \bibinfo{person}{Yujun Shen},
  \bibinfo{person}{Deli Zhao}, \bibinfo{person}{Jingren Zhou}, {and}
  \bibinfo{person}{Tieniu Tan}.} \bibinfo{year}{2023}\natexlab{}.
\newblock \showarticletitle{VideoFusion: Decomposed Diffusion Models for
  High-Quality Video Generation}. In \bibinfo{booktitle}{\emph{Proceedings of
  the IEEE/CVF Conference on Computer Vision and Pattern Recognition}}.
  \bibinfo{pages}{10209--10218}.
\newblock


\bibitem[Mahapatra and Kulkarni(2022)]%
        {mahapatra2022controllable}
\bibfield{author}{\bibinfo{person}{Aniruddha Mahapatra} {and}
  \bibinfo{person}{Kuldeep Kulkarni}.} \bibinfo{year}{2022}\natexlab{}.
\newblock \showarticletitle{Controllable Animation of Fluid Elements in Still
  Images}. In \bibinfo{booktitle}{\emph{Proceedings of the IEEE/CVF Conference
  on Computer Vision and Pattern Recognition}}. \bibinfo{pages}{3667--3676}.
\newblock


\bibitem[Maimon and Rokach(2005)]%
        {maimon2005data}
\bibfield{author}{\bibinfo{person}{Oded Maimon} {and} \bibinfo{person}{Lior
  Rokach}.} \bibinfo{year}{2005}\natexlab{}.
\newblock \showarticletitle{Data mining and knowledge discovery handbook}.
\newblock  (\bibinfo{year}{2005}).
\newblock


\bibitem[Peng et~al\mbox{.}(2022)]%
        {peng2022mpib}
\bibfield{author}{\bibinfo{person}{Juewen Peng}, \bibinfo{person}{Jianming
  Zhang}, \bibinfo{person}{Xianrui Luo}, \bibinfo{person}{Hao Lu},
  \bibinfo{person}{Ke Xian}, {and} \bibinfo{person}{Zhiguo Cao}.}
  \bibinfo{year}{2022}\natexlab{}.
\newblock \showarticletitle{Mpib: An mpi-based bokeh rendering framework for
  realistic partial occlusion effects}. In \bibinfo{booktitle}{\emph{European
  Conference on Computer Vision}}. Springer, \bibinfo{pages}{590--607}.
\newblock


\bibitem[Poole et~al\mbox{.}(2022)]%
        {poole2022dreamfusion}
\bibfield{author}{\bibinfo{person}{Ben Poole}, \bibinfo{person}{Ajay Jain},
  \bibinfo{person}{Jonathan~T Barron}, {and} \bibinfo{person}{Ben Mildenhall}.}
  \bibinfo{year}{2022}\natexlab{}.
\newblock \showarticletitle{Dreamfusion: Text-to-3d using 2d diffusion}.
\newblock \bibinfo{journal}{\emph{arXiv preprint arXiv:2209.14988}}
  (\bibinfo{year}{2022}).
\newblock


\bibitem[Ranftl et~al\mbox{.}(2021)]%
        {ranftl2021vision}
\bibfield{author}{\bibinfo{person}{Ren{\'e} Ranftl}, \bibinfo{person}{Alexey
  Bochkovskiy}, {and} \bibinfo{person}{Vladlen Koltun}.}
  \bibinfo{year}{2021}\natexlab{}.
\newblock \showarticletitle{Vision transformers for dense prediction}. In
  \bibinfo{booktitle}{\emph{Proceedings of the IEEE/CVF International
  Conference on Computer Vision}}. \bibinfo{pages}{12179--12188}.
\newblock


\bibitem[Ren and Wang(2022)]%
        {ren2022look}
\bibfield{author}{\bibinfo{person}{Xuanchi Ren} {and} \bibinfo{person}{Xiaolong
  Wang}.} \bibinfo{year}{2022}\natexlab{}.
\newblock \showarticletitle{Look outside the room: Synthesizing a consistent
  long-term 3d scene video from a single image}. In
  \bibinfo{booktitle}{\emph{Proceedings of the IEEE/CVF Conference on Computer
  Vision and Pattern Recognition}}. \bibinfo{pages}{3563--3573}.
\newblock


\bibitem[Ren et~al\mbox{.}(2020)]%
        {ren2020deep}
\bibfield{author}{\bibinfo{person}{Yurui Ren}, \bibinfo{person}{Xiaoming Yu},
  \bibinfo{person}{Junming Chen}, \bibinfo{person}{Thomas~H Li}, {and}
  \bibinfo{person}{Ge Li}.} \bibinfo{year}{2020}\natexlab{}.
\newblock \showarticletitle{Deep image spatial transformation for person image
  generation}. In \bibinfo{booktitle}{\emph{Proceedings of the IEEE/CVF
  Conference on Computer Vision and Pattern Recognition}}.
  \bibinfo{pages}{7690--7699}.
\newblock


\bibitem[Rockwell et~al\mbox{.}(2021)]%
        {rockwell2021pixelsynth}
\bibfield{author}{\bibinfo{person}{Chris Rockwell}, \bibinfo{person}{David~F
  Fouhey}, {and} \bibinfo{person}{Justin Johnson}.}
  \bibinfo{year}{2021}\natexlab{}.
\newblock \showarticletitle{Pixelsynth: Generating a 3d-consistent experience
  from a single image}. In \bibinfo{booktitle}{\emph{Proceedings of the
  IEEE/CVF International Conference on Computer Vision}}.
  \bibinfo{pages}{14104--14113}.
\newblock


\bibitem[Rombach et~al\mbox{.}(2022)]%
        {rombach2022high}
\bibfield{author}{\bibinfo{person}{Robin Rombach}, \bibinfo{person}{Andreas
  Blattmann}, \bibinfo{person}{Dominik Lorenz}, \bibinfo{person}{Patrick
  Esser}, {and} \bibinfo{person}{Bj{\"o}rn Ommer}.}
  \bibinfo{year}{2022}\natexlab{}.
\newblock \showarticletitle{High-resolution image synthesis with latent
  diffusion models}. In \bibinfo{booktitle}{\emph{Proceedings of the IEEE/CVF
  Conference on Computer Vision and Pattern Recognition}}.
  \bibinfo{pages}{10684--10695}.
\newblock


\bibitem[Saharia et~al\mbox{.}(2022)]%
        {saharia2022photorealistic}
\bibfield{author}{\bibinfo{person}{Chitwan Saharia}, \bibinfo{person}{William
  Chan}, \bibinfo{person}{Saurabh Saxena}, \bibinfo{person}{Lala Li},
  \bibinfo{person}{Jay Whang}, \bibinfo{person}{Emily~L Denton},
  \bibinfo{person}{Kamyar Ghasemipour}, \bibinfo{person}{Raphael
  Gontijo~Lopes}, \bibinfo{person}{Burcu Karagol~Ayan}, \bibinfo{person}{Tim
  Salimans}, {et~al\mbox{.}}} \bibinfo{year}{2022}\natexlab{}.
\newblock \showarticletitle{Photorealistic text-to-image diffusion models with
  deep language understanding}.
\newblock \bibinfo{journal}{\emph{Advances in Neural Information Processing
  Systems}}  \bibinfo{volume}{35} (\bibinfo{year}{2022}),
  \bibinfo{pages}{36479--36494}.
\newblock


\bibitem[Shade et~al\mbox{.}(1998)]%
        {shade1998layered}
\bibfield{author}{\bibinfo{person}{Jonathan Shade}, \bibinfo{person}{Steven
  Gortler}, \bibinfo{person}{Li-wei He}, {and} \bibinfo{person}{Richard
  Szeliski}.} \bibinfo{year}{1998}\natexlab{}.
\newblock \showarticletitle{Layered depth images}. In
  \bibinfo{booktitle}{\emph{Proceedings of the 25th annual conference on
  Computer graphics and interactive techniques}}. \bibinfo{pages}{231--242}.
\newblock


\bibitem[Shih et~al\mbox{.}(2020)]%
        {shih20203d}
\bibfield{author}{\bibinfo{person}{Meng-Li Shih}, \bibinfo{person}{Shih-Yang
  Su}, \bibinfo{person}{Johannes Kopf}, {and} \bibinfo{person}{Jia-Bin Huang}.}
  \bibinfo{year}{2020}\natexlab{}.
\newblock \showarticletitle{3d photography using context-aware layered depth
  inpainting}. In \bibinfo{booktitle}{\emph{Proceedings of the IEEE/CVF
  Conference on Computer Vision and Pattern Recognition}}.
  \bibinfo{pages}{8028--8038}.
\newblock


\bibitem[Siarohin et~al\mbox{.}(2019)]%
        {siarohin2019first}
\bibfield{author}{\bibinfo{person}{Aliaksandr Siarohin},
  \bibinfo{person}{St{\'e}phane Lathuili{\`e}re}, \bibinfo{person}{Sergey
  Tulyakov}, \bibinfo{person}{Elisa Ricci}, {and} \bibinfo{person}{Nicu Sebe}.}
  \bibinfo{year}{2019}\natexlab{}.
\newblock \showarticletitle{First order motion model for image animation}.
\newblock \bibinfo{journal}{\emph{Advances in Neural Information Processing
  Systems}}  \bibinfo{volume}{32} (\bibinfo{year}{2019}).
\newblock


\bibitem[Siarohin et~al\mbox{.}(2021)]%
        {siarohin2021motion}
\bibfield{author}{\bibinfo{person}{Aliaksandr Siarohin},
  \bibinfo{person}{Oliver~J Woodford}, \bibinfo{person}{Jian Ren},
  \bibinfo{person}{Menglei Chai}, {and} \bibinfo{person}{Sergey Tulyakov}.}
  \bibinfo{year}{2021}\natexlab{}.
\newblock \showarticletitle{Motion representations for articulated animation}.
  In \bibinfo{booktitle}{\emph{Proceedings of the IEEE/CVF Conference on
  Computer Vision and Pattern Recognition}}. \bibinfo{pages}{13653--13662}.
\newblock


\bibitem[Singer et~al\mbox{.}(2022)]%
        {singer2022make}
\bibfield{author}{\bibinfo{person}{Uriel Singer}, \bibinfo{person}{Adam
  Polyak}, \bibinfo{person}{Thomas Hayes}, \bibinfo{person}{Xi Yin},
  \bibinfo{person}{Jie An}, \bibinfo{person}{Songyang Zhang},
  \bibinfo{person}{Qiyuan Hu}, \bibinfo{person}{Harry Yang},
  \bibinfo{person}{Oron Ashual}, \bibinfo{person}{Oran Gafni}, {et~al\mbox{.}}}
  \bibinfo{year}{2022}\natexlab{}.
\newblock \showarticletitle{Make-a-video: Text-to-video generation without
  text-video data}.
\newblock \bibinfo{journal}{\emph{arXiv preprint arXiv:2209.14792}}
  (\bibinfo{year}{2022}).
\newblock


\bibitem[Skorokhodov et~al\mbox{.}(2021)]%
        {skorokhodov2021aligning}
\bibfield{author}{\bibinfo{person}{Ivan Skorokhodov}, \bibinfo{person}{Grigorii
  Sotnikov}, {and} \bibinfo{person}{Mohamed Elhoseiny}.}
  \bibinfo{year}{2021}\natexlab{}.
\newblock \showarticletitle{Aligning latent and image spaces to connect the
  unconnectable}. In \bibinfo{booktitle}{\emph{Proceedings of the IEEE/CVF
  International Conference on Computer Vision}}. \bibinfo{pages}{14144--14153}.
\newblock


\bibitem[Wang et~al\mbox{.}(2022)]%
        {wang20223d}
\bibfield{author}{\bibinfo{person}{Qianqian Wang}, \bibinfo{person}{Zhengqi
  Li}, \bibinfo{person}{David Salesin}, \bibinfo{person}{Noah Snavely},
  \bibinfo{person}{Brian Curless}, {and} \bibinfo{person}{Janne Kontkanen}.}
  \bibinfo{year}{2022}\natexlab{}.
\newblock \showarticletitle{3D moments from near-duplicate photos}. In
  \bibinfo{booktitle}{\emph{Proceedings of the IEEE/CVF Conference on Computer
  Vision and Pattern Recognition}}. \bibinfo{pages}{3906--3915}.
\newblock


\bibitem[Wiles et~al\mbox{.}(2020)]%
        {wiles2020synsin}
\bibfield{author}{\bibinfo{person}{Olivia Wiles}, \bibinfo{person}{Georgia
  Gkioxari}, \bibinfo{person}{Richard Szeliski}, {and} \bibinfo{person}{Justin
  Johnson}.} \bibinfo{year}{2020}\natexlab{}.
\newblock \showarticletitle{Synsin: End-to-end view synthesis from a single
  image}. In \bibinfo{booktitle}{\emph{Proceedings of the IEEE/CVF Conference
  on Computer Vision and Pattern Recognition}}. \bibinfo{pages}{7467--7477}.
\newblock


\bibitem[Zhang et~al\mbox{.}(2018)]%
        {zhang2018unreasonable}
\bibfield{author}{\bibinfo{person}{Richard Zhang}, \bibinfo{person}{Phillip
  Isola}, \bibinfo{person}{Alexei~A Efros}, \bibinfo{person}{Eli Shechtman},
  {and} \bibinfo{person}{Oliver Wang}.} \bibinfo{year}{2018}\natexlab{}.
\newblock \showarticletitle{The unreasonable effectiveness of deep features as
  a perceptual metric}. In \bibinfo{booktitle}{\emph{Proceedings of the IEEE
  conference on computer vision and pattern recognition}}.
  \bibinfo{pages}{586--595}.
\newblock


\end{thebibliography}










\end{document}